\documentclass[runningheads]{llncs}

\usepackage{eccv}

\usepackage{eccvabbrv}

\usepackage{graphicx}
\usepackage{booktabs}

\usepackage[accsupp]{axessibility}  

\usepackage[pagebackref,breaklinks,colorlinks,citecolor=eccvblue]{hyperref}

\usepackage{orcidlink}

\newcommand{\method}{DRDC\xspace}
\newcommand{\jinxiang}[1]{\textcolor{black}{#1}}
\newcommand{\sylvia}[1]{\textcolor{black}{#1}}

\newcommand{\beforesectionvspace}{\vspace{-0.2cm}}
\newcommand{\aftersectionvspace}{\vspace{-0.2cm}}
\newcommand{\subsectionvspace}{\vspace{-0.1cm}}

\usepackage{algorithm}
\usepackage{algorithmic}
\usepackage{newfloat}
\usepackage{listings}
\usepackage{dsfont,pifont}
\usepackage{amsmath}
\usepackage{amssymb}
\usepackage{graphicx}
\usepackage{booktabs}
\usepackage{soul}
\usepackage{multirow}
\usepackage{tikz}
\usepackage{pgfplots}
\usepackage{xcolor}
\usepackage{xspace}
\usepackage{mathtools}
\usepackage{amsfonts}
\usepackage{array}

\begin{document}
	
	\title{Enhancing Multi-Class Anomaly Detection via Diffusion Refinement with Dual Conditioning} 
	
	
	\author{Jiawei Zhan\inst{1} \and
		Jinxiang Lai\inst{1} \and
		Bin-Bin Gao\inst{1} \and
  Jun Liu\inst{1} \and
  Xiaochen Chen\inst{1} \and
  Chengjie Wang\inst{1}
  }
	
	\authorrunning{J. Zhan et al.}
	\titlerunning{Enhancing Multi-Class Anomaly Detection via DRDC}

	\institute{Tencent Youtu Lab, Shenzhen, CN \\
		\email{\{gavynzhan,jinxianglai,danylgao,juliusliu,husonchen,jasoncjwang\}@tencent.com}}
	
	\maketitle

	\begin{abstract}
		\vspace{-0.3cm}
		
		Anomaly detection, the technique of identifying abnormal samples using only normal samples, has attracted widespread interest in industry. Existing one-model-per-category methods often struggle with limited generalization capabilities due to their focus on a single category, and can fail when encountering variations in product. Recent feature reconstruction methods, as representatives in one-model–all-categories schemes, face challenges including reconstructing anomalous samples and blurry reconstructions.
		In this paper, we creatively combine a diffusion model and a transformer for multi-class anomaly detection. This approach leverages diffusion to obtain high-frequency information for refinement, greatly alleviating the blurry reconstruction problem while maintaining the sampling efficiency of the reverse diffusion process. 
		The task is transformed into image inpainting to disconnect the input-output correlation, thereby mitigating the ``identical shortcuts'' problem and avoiding the model from reconstructing anomalous samples.
		Besides, we introduce category-awareness using dual conditioning to ensure the accuracy of prediction and reconstruction in the reverse diffusion process, preventing excessive deviation from the target category, thus effectively enabling multi-class anomaly detection. 
		Futhermore, Spatio-temporal fusion is also employed to fuse heatmaps predicted at different timesteps and scales, enhancing the performance of multi-class anomaly detection.
		Extensive experiments on benchmark datasets demonstrate the superior performance and exceptional multi-class anomaly detection capabilities of our proposed method compared to others.
		\vspace{-0.2cm}
		
	\end{abstract}
	
	\vspace{-0.4cm}
	
	\beforesectionvspace
	\section{Introduction}
	\aftersectionvspace
	\label{sec:intro}
	Anomaly detection, designed to trained on normal samples to identify abnormal images and locate abnormal regions, has become an indispensable tool in various fields including industrial defect detection~\cite{mvtec}, medical image analysis~\cite{medical}, and video surveillance~\cite{synthesizecompare}.
	In industrial settings, anomaly detection is particularly challenging due to the scarcity of abnormal samples and the wide range of anomalies, which can range from subtle changes to large structural defects. 
	Although existing methods~\cite{us,padim,cutpaste,fcdd,psvdd,draem} offer acceptable performance, the practicality of their one-model–per-category scheme is debatable. 
	These methods can easily fail and have limited generalization performance when trained on categories with large deviations~\cite{uniad}. 
	Conversely, the one-model-all-categories scheme, which uses a unified framework to detect anomalies across various categories, can operate more effectively and reduce deployment costs in such scenarios.

	As representatives of multi-class anomaly detection, reconstruction-based anomaly detection methods \cite{ganomaly,l2ae_ssimae,vevae,featurerecon,oldgold} have recently garnered increased interest. The generalization performance of such models allows them to be seamlessly ported into a unified model for multiple classes without compromising performance. However, these methods are not without their challenges. Firstly, they operate under the assumption that the model will only reconstruct normal samples after being trained on normal samples, but the model has the potential to reconstruct anomalous samples, resulting in low reconstruction errors that fail to identify anomalous regions. Secondly, these methods involve processes like downsampling, which leads to the problem of blurred outputs and may cause the model to potentially overlook thin defects, negatively impacting the performance of anomaly detection.
	
	To alleviate these problems, we propose a novel framework for unsupervised multi-class anomaly detection named Diffusion Refinement with Dual Conditioning (\method).
	\ul{Why choose diffusion?}
	\ding{182} Several typical unsupervised anomaly detection methods using generative models, such as GANs~\cite{gan,ganomaly,ocgan,oldgold} and VAEs~\cite{pinaya2021unsupervised,NIPS2017_7a98af17}, still suffer from potential resolution loss due to pooling and strided convolutions. In addition, their performance is unsatisfactory, and the training process can become unstable on larger samples~\cite{miyato2018spectral,brock2017neural,brock2018large}. Since diffusion models offer excellent inductive biases for spatial data, the heavy spatial downsampling of related generative models is not required~\cite{ddpm,ldm}. Besides, the spatial dimensions of intermediate variables remain constant at every timestep, resulting in less loss of spatial information.
	\ding{183} You and Cui~\cite{uniad} provide insight into the nature of ``identical shortcuts''; that is the the characteristic of models to map inputs directly to outputs via shortcuts, resulting in the tendency to reconstruct anomalous samples when the inputs are abnormal. The ``identity shortcut'' problem can be mitigated, however, by inheriting the diffusion property that the back-end network uses to predict noise rather than predict images. 
	
	\ul{How to exploit diffusion?} 
	\ding{182} Our method alleviates the blurry reconstruction problem by applying high-resolution refinement to the feature reconstruction-based model. The refinement is achieved by applying the diffusion model to an inpainting task that isolates the correlation with known and masked regions by making predictions, preventing the model from mapping the input region directly to the output (i.e., potential leakage of ``identical shortcuts'') and thus further mitigating the tendency to reconstruct anomalous samples. Moreover, due to the slow sampling speed of diffusion, direct reconstruction by diffusion either requires an unacceptable cost for inference time or a compromise in reconstruction accuracy. In contrast, we use the diffusion method to reconstruct the high-frequency part, which not only allows faster sampling, but also efficiently utilizes the underlying transformer-based capabilities.
	\ding{183} The one-model–all-categories scheme poses a challenge when dealing with different anomaly definitions across categories, as what is considered anomalous in one category may be regarded as normal in another. Direct incorporation of the original diffusion can lead to uncontrollable randomness in the reconstructed images. While this wouldn't affect the one-model–per-category scheme as it only utilizes one category for training, however, it can lead to reconstructing samples of different semantic categories in the ``one-model-all-categories'' scenario, severely affecting performance. To achieve category-awareness to handle different definitions for multi-class anomaly detection, we introduce dual conditioning to ensure that samples can be accurately reconstructed in the direction of the expected semantic category.
	\ding{184} Moreover, we propose a Spatio-temporal fusion module to perform a smoother fusion of heatmaps predicted at different timesteps and scales, which helps to mitigate the accumulative errors that can arise from relying solely on the final prediction or the bias of limited information at each timestep.
	
	We conduct extensive experiments on the challenging MVTec-AD~\cite{mvtec} and BeanTechAD~\cite{vtadl} to demonstrate the effectiveness of our method under the unified setting. Note that our proposed \method achieves state-of-the-art 98.5 image-wise AUROC and 98.1 pixel-wise AUROC for multi-class anomaly detection on MVTec-AD.
	Our main contributions include:
	\begin{itemize}
		\item We determine the adverse effects of low spatial resolution from reconstruction methods. As a result, we propose a novel multi-class anomaly detection framework with diffusion refinement as a solution.
		\item Unlike the previous method of using diffusion in a trivial way, we avoid ``identical shortcuts'' by using diffusion for inpainting, while exploiting only the high-frequency part to refine the low-resolution heatmap, thus improving the sampling speed and taking advantage of diffusion model.
		\item To ensure accuracy and to accommodate the setting of multi-class anomaly detection, we introduce the technique of dual conditioning and propose the module of Spatio-temporal fusion, which enables a smooth fusion of heatmaps predicted at different timesteps and scales.
		\item Extensive experiments on MVTec-AD and BeanTechAD demonstrate the superiority of our framework over existing alternatives under the unified task setting.
	\end{itemize}
	
	\beforesectionvspace
	\section{Related Work}
	\aftersectionvspace
	
	\textbf{Anomaly Detection.}
	Existing unsupervised anomaly detection techniques can be classified into classical methods \cite{ocsvm,fcdd}, distribution-based methods \cite{padim,modelingdistribution}, and reconstruction-based methods \cite{l2ae_ssimae,vevae,adtr}.
	These one-model-per-category methods, however, are not well-suited when extending to multiple categories, as they hinge on a single hyper-sphere or a single distribution, which may not adequately capture the varying feature representation among categories.
	Recently, multi-class unified methods have received attention due to their practicality~\cite{regad,metaformer,uniad}. 
	The most relevant UniAD~\cite{uniad} constructs a unified model for multiple categories via feature reconstruction, but its low spatial resolution (only 14$\times$14 spatial resolution) of input features hinders its ability to precisely locate defects. 
	Consequently, we introduce a diffusion-based refinement strategy for multi-class anomaly detection. This not only circumvents the issue of blurry reconstruction but also ensures superior anomaly detection performance, surpassing current state-of-the-art methods.
	
	\noindent\textbf{Diffusion Models.}
	Recently, DDPMs have been in the spotlight for their ability on image synthesis~\cite{diffusionbeatgans}, with superior mode coverage~\cite{xiao2022tackling} over generative adversarial networks (GANs) and variational autoencoders (VAEs)~\cite{9555209}. 
	Although there have been several endeavors in the anomaly detection domain~\cite{AnoDDPM,wolleb2022diffusion,mousakhan2023anomaly}, they primarily focus on medical image detection or follows one-model-per-category scheme, which presents application scenarios different from our multi-class anomaly detection. In addition, they have simply utilized diffusion models that offer mediocre performance and require a greater number of iteration steps.
	DiAD~\cite{diad} attempts to implement anomaly detection using LDM~\cite{ldm}, but still suffers from the problem of blurring due to his reconstruction of embedding.
	In contrast, we maintain the high resolution of the intermediate variables and exploit only high-frequency information to increase the sampling speed while converting to inpainting tasks to avoid ``identical shortcuts'', thereby enhancing the usability of diffusion for anomaly detection.
	
	\beforesectionvspace
	\section{Preliminaries}
	\aftersectionvspace

	\begin{table}[!b]
		\setlength\tabcolsep{5.0pt}
		\centering
		\footnotesize
		\vspace{-0.3cm}
		\caption{Anomaly localization results of scaled ground-truth mask with AUROC metric on MVTec-AD. The result is the average of 15 categories.}
			\begin{tabular}{l|l|l|l|l|l}
				\toprule
				\textbf{Scaling Size} & 1 & 1/2 & 1/4 & 1/8 & 1/16 \\   
				\midrule
				\textbf{Pixel-wise AUROC} & 100.00 & 99.99 & 99.93 & 99.27 & 96.05 \\
				\bottomrule

			\end{tabular}
		\vspace{-0.3cm}
      \label{tab:discussion}
	\end{table}

	\subsection{Base Model with Feature Reconstruction}
	For unsupervised multi-class anomaly detection, a typical attempt is to use feature reconstruction to construct the base model, and discriminate the anomaly via measuring the reconstruction error between the reconstructed sample and the original one.
	\sylvia{As shown in \cref{fig:overview}, the base model utilizes a transformer network to generate reconstructed feature for any input feature.}
	\jinxiang{Formally, the base model is optimized by the Mean Squared Error (MSE) loss,
		\begin{equation}
			\mathcal{L}_{\text{feat}}=\frac{1}{H_{\text{feat}}\times W_{\text{feat}}}{||F_{\text{in}}-F_{\text{out}}||}^2_2,
		\end{equation}
		where, $F_{\text{in}} \in \mathbb {R}^{C_{\text{feat}}\times H_{\text{feat}}\times W_{\text{feat}}}$ is the input feature extracted from the original image by the pre-trained backbone network with fixed weights, and $F_{\text{out}}\in\mathbb{R}^{C_{\text{feat}}\times H_{\text{feat}}\times W_{\text{feat}}}$ is the recovered output feature generated by the transformer network.
		Note that $C_{\text{feat}}$, $H_{\text{feat}}$ and $W_{\text{feat}}$ represent the channel number, height, and width of the feature, respectively.}
	
	
	
	\jinxiang{The anomaly result predicted by the base model is represented as a base heatmap $\mathcal{H}_{\text{base}} \in\mathbb{R}^{H_{\text{feat}}\times W_{\text{feat}}}$, which is calculated as the L2 norm of the reconstruction differences,}
	\begin{equation}
		\mathcal{H}_{\text{base}}={||F_{\text{in}}-F_{\text{out}}||}_2.
	\end{equation}
	
	\subsectionvspace
	\subsection{Blurry Problem of Base Model}
	As previously stated, due to the limited spatial size of the features, with a resolution of merely 14$\times$14, the final representation fails to accurately localize defects when the original resolution heatmap of 224$\times$224 is obtained through direct upsampling, let alone for higher resolution. To illustrate this phenomenon, we downsample the ground-truth mask directly to a lower resolution, and then upsample it back to the original size. This process illustrates the significant performance loss that even the ground-truth mask would experience under such downsampling treatment. For simplicity of illustration, we give the final pixel-wise AUROC averaged over 15 categories on MVTec-AD~\cite{mvtec}.
	
	As revealed in \Cref{tab:discussion},
	the pixel-wise AUROC impact is not substantial when the size is reduced by half. However, when reduced to 1/16, the pixel-wise AUROC performance declines more significantly. Note that these results cannot be directly regarded as the upper bound, since the data-type of model outputs is different from that of the ground-truth mask (floating-point scores can contain more information vs. binary masks can easily lose thin defects during downsampling).
	However, such results tell the limitations of small-size reconstruction. For this reason, we will perform a refinement to improve the anomaly heatmaps.
	
	\subsectionvspace
	\subsection{Diffusion Models}
	As outlined in~\cite{diffusion,ddpm}, diffusion models consist of a forward diffusion process and a reverse denoising process. The forward process is a Markov process that transforms an image $\textbf{x}_0$ into white Gaussian noise $\textbf{x}_T\sim\mathcal{N}(\textbf{0},\textbf{I})$ over a series of $T$ timesteps. 
	Each step is defined as,
	\begin{equation}\label{eq:xprevtoxcurr}
		q(\textbf{x}_{t}|\textbf{x}_{t-1})=\mathcal{N}(\textbf{x}_{t};\sqrt{1-\beta_t}\textbf{x}_{t-1},\beta_t\textbf{I}),
	\end{equation}
	By utilizing the independence property of the noise added at each step, as described in \cref{eq:xprevtoxcurr}, we can calculate the total noise variance as $\Bar{\alpha}_t=\prod^t_{s=1}(1-\beta_s)$, thereby marginalizing the forward process at each step,
	\begin{equation}\label{eq:rewrite}
		q(\textbf{x}_t|\textbf{x}_0)=\mathcal{N}(\textbf{x}_t;\sqrt{\Bar{\alpha}_t}\textbf{x}_0,(1-\Bar{\alpha}_t)\textbf{I}).
	\end{equation}
	In the reverse denoising process, $\textbf{x}_{t-1}$ is iteratively reconstructed from $\textbf{x}_{t}$. The process is modeled by a U-Net~\cite{unet} network $z_\theta$ that predicts the distribution,
	\begin{equation}\label{eq:xcurrtoxprev}
		p_\theta(\textbf{x}_{t-1}|\textbf{x}_{t})=\mathcal{N}(\textbf{x}_{t-1};\mu_\theta,\Sigma_\theta).
	\end{equation}
	where $\mu_\theta=\frac{1}{\sqrt{\alpha_t}}(\textbf{x}_t-\frac{\beta_t}{\sqrt{1-\Bar{\alpha}_t}}z_\theta)$, $\Sigma_\theta=\sigma_t^2\textbf{I}$, $\sigma_t$ is a fixed constant related to the variance schedule and $\theta$ is the learnable parameter which could be optimized as,
	\begin{equation}
		\label{eq:final_loss}
		\mathop{\min}_{\theta}\mathbb{E}_{t,\textbf{x}_0\sim q(\textbf{x}_0),\epsilon\sim\mathcal{N}(\textbf{0},\textbf{I})}\left[{||\epsilon-z_\theta(\textbf{x}_t,t)||}^2_2\right].
	\end{equation}

	\begin{figure*}[!t]
		\centering
		\includegraphics[width=0.90\linewidth]{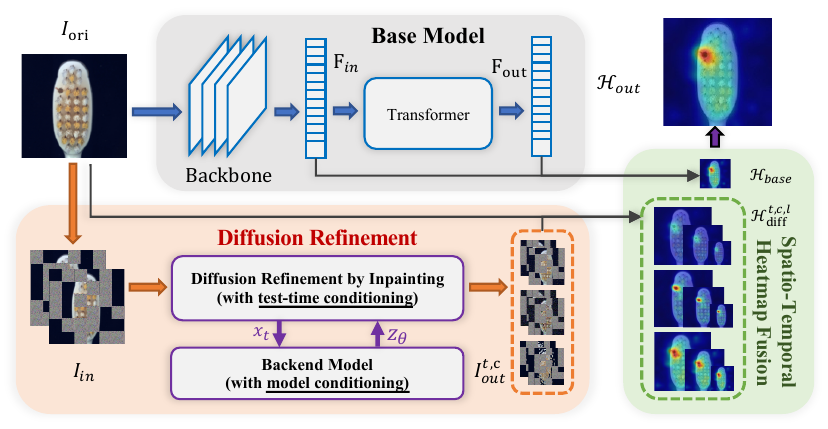}
		\vspace{-0.1cm}
		
		\caption{The overview of our proposed framework consists of the Base Model, Diffusion Refinement, and Spatio-temporal Heatmap Fusion. Samples are fed into both the base model and the diffusion refinement model. The base model produces a low-resolution base heatmap, while the diffusion refinement model generates high-resolution high-frequency correction heatmaps. Finally, spatio-temporal fusion is employed to obtain the final anomaly map.}
		\label{fig:overview}
		\vspace{-0.3cm}
		
	\end{figure*}
	
	\beforesectionvspace
	\section{Methodology}
	\aftersectionvspace
	
	\subsection{Our \method Framework}
	As depicted in \cref{fig:overview}, our framework primarily consists of three components, each with distinct functionalities. 
	The base model, supported by a Transformer, reconstructs features to create a low-resolution base heatmap. 
	The diffusion refinement model, backed by U-Net~\cite{unet} and guided by a dual conditioning strategy, ensures class-awareness and improves the accuracy of the reverse diffusion process when generating multiple high-resolution correction heatmaps that contain high-frequency information.
	Lastly, the spatio-temporal fusion module merges the generated heatmaps from various timesteps, multiple scales, and different inpainting specifications to produce the final output.
	
	
	\subsectionvspace
	\subsection{Diffusion Refinement by Inpainting}
	While the base model can produce acceptable heatmap, its limited resolution restricts its ability to accurately localize subtle defects, resulting in missed detections. 
	In particular, categories such as screws and capsules, which contain thin strips of defects, are constrained by the low spatial resolution of the feature reconstruction. 
	To tackle this issue, we employ a diffusion model for image inpainting, which reconstructs the masked region by introducing noise and gradually reducing it through a reverse diffusion process. Importantly, we perform inpainting at the same resolution as the original input, allowing us to preserve detailed spatial information. Since the network is trained on normal samples, the recovered region represents what it would appear like under normal conditions.

	
	For clarity, let's define $\textbf{x}_0\!=\!I_{\text{in}}$, which serves as the input to this module.
	The input image $\textbf{x}_0$ is created by replacing a set of pixels with Gaussian noise. 
	Specifically, we first divide the input image into $G\!=\!\frac{H_{\text{img}}}{c}\!\times\!\frac{W_{\text{img}}}{c}$ grids, 
	where $H_{\text{img}}$ and $W_{\text{img}}$ are the image's height and width, and $c\!\in\!\{c_i\}_{i=1}^{n_c}$ is a factor of both $H_{\text{img}}$ and $W_{\text{img}}$.
	These grids are then randomly and uniformly partitioned into $n_s$ disjoint sets $\mathbb{S}_i$, each containing $\frac{G}{n_s}$ grid elements, with each element being a $c\times c$ pixel square.
	${\mathbb{M}_i}\!\in\!\mathbb{M}$ is the corresponding binary mask that contains zeros in regions belonging to $\mathbb{S}_i$ and ones in other regions.
	During inference, ${\mathbb{M}_i}$ is used to replace the regions in the image that belong to $\mathbb{S}_i$ with Gaussian noise, transforming $I_{\text{ori}}$ into $I_{\text{in}}$.
	Then, the regions replaced with noise are gradually reconstructed back to $\Tilde{\textbf{x}}_0$ after being processed forward as $\textbf{x}_t$.
	
	To leverage different information at different moments, we can apply a set of different timesteps $t\!\in\!\{t_i\}_{i=1}^{n_t}$. For each $t$, we can predict the masked region at timestep $0$ using a reverse diffusion process as follows according to \cref{eq:rewrite},
	\begin{equation}
		\Tilde{\textbf{x}}_0^t=\frac{1}{\sqrt{\Bar{\alpha}_t}}(\textbf{x}_t-\sqrt{1-\Bar{\alpha}_t}z_\theta).
	\end{equation}
	
	Combining the unknown part of each disjoint set $\mathbb{S}_i$ using its corresponding binary mask $\mathbb{M}_i$, we can get the prediction $I_{\text{out}}^{t,c}$ of grid size $c$ at timestep $t$,
	\begin{equation}\label{eq:I_out}
		I_{\text{out}}^{t,c}=\frac{1}{n_s}\sum_{{\mathbb{M}_i}\in\mathbb{M}}{(1-{\mathbb{M}_i})\odot\Tilde{\textbf{x}}_0^t}.
	\end{equation}
	As a result, heatmap $\mathcal{H}_{\text{diff}}^{t,c}$ can be obtained as follows,
	\begin{equation}\label{eq:H_diff}
		\mathcal{H}_{\text{diff}}^{t,c}={||I_{\text{ori}}-I_{\text{out}}^{t,c}||}_2\in\mathbb{R}^{H_{\text{img}}\times W_{\text{img}}}.
	\end{equation}

	As per prior studies~\cite{ilvr,sdedit}, in the forward process, the original data progressively sheds information, with high-frequency details being lost before the low-frequency ones. Conversely, in the reverse process, information is gradually recovered from pure noise, with low-frequency details being obtained before the high-frequency ones.
	Drawing from these observations, 
	we execute only a few timesteps $\{t\}$ during the early stages of the diffusion, enabling high-frequency information retention and a faster reverse process.
	
	
	\subsectionvspace
	\subsection{Dual Conditioning on the Known Region}
	
	To perform multi-class unsupervised anomaly detection, it is necessary to accurately determine the category of a given sample. The base model encodes category information through an additional query embedding, while we utilize diffusion to implicitly encode categories as conditions without the need for an explicit classifier or explicit category supervision. The introduction of model conditioning and test-time conditioning ensures that the model does not misclassify similar categories, preventing it from incorrectly filling the missing region with the wrong semantic category.
	
	\noindent\textbf{Model Conditioning}:
	Recall that the neural network $z_\theta$ is trained to estimate $\epsilon$ for any given noisy image $\textbf{x}_t$.
	To improve the efficiency and accuracy of the neural network, we setup category-awareness into the model by adding conditions. Specifically, we condition the model on the input image $I_{\text{in}}$ [with noise and mask], allowing it to learn category-specific noise distributions and accurately reconstruct the target noise during the reverse diffusion process. As we aim to accelerate the inference process by limiting the number of timesteps, it is important to ensure the accuracy of the predictions at each single timestep. Presetting conditions during model training can effectively enable the model to be aware of the category and perform implicit classification, thereby accurately predicting noise. As a result, we extend \cref{eq:final_loss} as follows,
	\begin{equation}
		\mathcal{L}_{\text{diff}}=\mathbb{E}_{t,\textbf{x}_0,I_{\text{in}},\epsilon}\left[{||\epsilon-z_\theta(\textbf{x}_t,I_{\text{in}},t)||}^2_2\right],
	\end{equation}
	where $\textbf{x}_t$ is the sample at timestep $t$, $I_{\text{in}}$ is our image condition. In practical implementation, $I_{\text{in}}$ and $I_t$ are concatenated at the channel level and proceed to subsequent modules for noise prediction.
	Note that we do not direct exploit our binary inpainting mask as in previous work~\cite{yu2019free,riad}. Instead, we fill the masked region of original images with Gaussian noise to meet the requirements of the diffusion process. The effectiveness for the model conditioning will be demonstrated in the ablation study section.

	\noindent\textbf{Test-time Conditioning}:
 	The conditioning at the test-time ensures that the model can guide the inpainting process of reverse diffusion, preventing excessive deviations from the original category. The crux of this conditioning is preserving a portion of the original category/sample information from the spatial dimension during inference, enabling the model to adjust and reconstruct regions using spatial information. This approach differs from model conditioning, as it prioritizes spatial structure alignment over mere category semantics matching.
	Drawing inspiration from~\cite{repaint}, we denote the ground-truth image as $\textbf{x}$, the known pixels can be ${\mathbb{M}_i}\odot\textbf{x}$ and the unknown pixels can be $(1-{\mathbb{M}_i})\odot\textbf{x}$.
	
	Given that every reverse step (from $\textbf{x}_t$ to $\textbf{x}_{t-1}$) in \cref{eq:xcurrtoxprev} depends solely on $\textbf{x}_t$, we can alter the known regions and thereby insert information into the reverse process to control its noise reduction direction as long as we keep the correct properties of the corresponding distribution. 
	Since the forward process in \cref{eq:xprevtoxcurr} is characterized by a Markov Chain of added Gaussian noise, we can obtain the image $\textbf{x}_{t-1}$ for the known region at any given time using \cref{eq:rewrite},
	\begin{equation}
		\textbf{x}^{\text{known}}_{t-1}\sim\mathcal{N}(\sqrt{\Bar{\alpha}_{t-1}}\textbf{x}_0,(1-\Bar{\alpha}_{t-1})\textbf{I}).
	\end{equation}
	
	This allows us to generate the known regions ${\mathbb{M}_i}\odot \textbf{x}_{t-1}$ at any timestep.
	Meanwhile, we can predict the image $\textbf{x}_{t-1}$ for the unknown region using \cref{eq:xcurrtoxprev},
	\begin{equation}
		\textbf{x}^{\text{unknown}}_{t-1}\sim\mathcal{N}(\mu_\theta(\textbf{x}_t,I_{\text{in}},t),\Sigma_\theta(\textbf{x}_t,I_{\text{in}},t)).
	\end{equation}
	
	Therefore, we achieve the following expression for one reverse step in our approach instead of \cref{eq:xcurrtoxprev},
	\begin{equation}
		\textbf{x}_{t-1}={\mathbb{M}_i}\odot\textbf{x}^{\text{known}}_{t-1}+(1-{\mathbb{M}_i})\odot \textbf{x}^{\text{unknown}}_{t-1}.
	\end{equation}
	
	In short, $\textbf{x}^{\text{known}}_{t-1}$ is generated using the known pixels in the given image $\textbf{x}_0$, while $\textbf{x}^{\text{unknown}}_{t-1}$ is sampled from the model given the previous iteration $\textbf{x}_t$. The results are then combined to the new sample $\textbf{x}_{t-1}$ using the mask.
	
	\subsectionvspace
	\subsection{Spatio-temporal Heatmap Fusion}
	
	The Spatio-temporal fusion module is aimed to fuse the low-resolution heatmaps from the base model and high-resolution heatmaps from the diffusion model.
	
	
	Recall that each step of the reverse diffusion process is able to predict a noise and to derive its corresponding predicted original image. 
	However, each prediction corresponding to current timestep leads to bias due to the limited information available. 
	In contrast, using only the final predictions of the reverse diffusion leads to accumulative errors~\cite{vqdiffusion,Schmidt2019GeneralizationIG}. 
	In addition, several studies~\cite{mkd,us,csflow} have shown that performing fusion at different spatial scales can be helpful for anomaly detection.
	For the diffusion reconstructed heatmap, since anomalies tend to occupy larger spatially connected regions, the reconstruction error can be aggregated over a larger region for more accurate anomaly detection.
	All things considered, we propose a Spatio-temporal fusion module to fuse various timesteps and spatial scales.
	For this purpose, we further extend \cref{eq:H_diff} with a scaled $l\!\in\!\{l_i\}_{i=1}^{n_l}$.
	Thus, the anomaly score map for each scale $l$ is calculated by downsampling the orginal image $I_{\text{ori}}$ and output $I_{\text{out}}$ to the scale $\frac{1}{l}$ and then upsampling to the original resolution,
	\begin{equation}
		\mathcal{H}_{\text{diff}}^{t,c,l}=U(||D(I_{\text{ori}},\frac{1}{l})-D(I_{\text{out}}^{t,c},\frac{1}{l})||_2,l)\in\mathbb{R}^{H_{\text{img}}\times W_{\text{img}}},
	\end{equation}
	where $U(\cdot)$, $D(\cdot)$ represent the upsampling and downsampling operation, respectively. Due to the low spatial resolution of the feature reconstruction, multi-scale fusion of the base heatmap does not provide extra performance gains, as will be demonstrated in the ablation study. 
	The Spatio-temporal heatmap $\mathcal{H}_{\text{diff}}^{\text{ST}}$ is then computed as the per-pixel average of all generated heatmaps with varying timesteps, multiple scales, and different inpainting specifications,
	\begin{equation}
		\mathcal{H}_{\text{diff}}^{\text{ST}}=\frac{1}{{n_t}{n_c}{n_l}}\sum_{i=1}^{{n_t}}\sum_{j=1}^{{n_c}}\sum_{k=1}^{{n_l}}\mathcal{H}_{\text{diff}}^{t_i,c_j,l_k}\in\mathbb{R}^{H_{\text{img}}\times W_{\text{img}}}.
	\end{equation}

	
	
	Since the predicted $I_{\text{out}}^{t,c}$ are obtained through respective reverse diffusion process at different timesteps, they contain some noise along with the limited information.
	As a result, the heatmaps can be smoothed to obtain a further optimized representation $\mathcal{H}_{\text{diff}}^{\text{SST}}$, which is post-processed by a mean-filter convolution,
	\begin{equation}
		\mathcal{H}_{\text{diff}}^{\text{SST}}=\mathcal{H}_{\text{diff}}^{\text{ST}}*f_{m\times m}\in\mathbb{R}^{H_{\text{img}}\times W_{\text{img}}},
	\end{equation}
	where $f_{m\times m}$ is the mean filter of size ($m\!\times\!m$) used for smoothing, $*$ is the convolution operation.
	In addition, we set a hyper-parameter $\gamma$ to adjust the weight on heatmaps,
	\begin{equation}
		\mathcal{H}_{\text{out}}=(1-\gamma)\frac{\mathcal{H}_{\text{base}}}{C_{\text{feat}}}+\gamma\frac{\mathcal{H}_{\text{diff}}^{\text{SST}}}{C_{\text{img}}}.
	\end{equation}
	
	our aim is to detect whether an image contains anomalous regions, which can be obtained by taking the maximum value of the averagely pooled $\mathcal{H}_{\text{out}}$.

	\begin{table*}[t]
	\setlength\tabcolsep{2.0pt}
	\centering
	\footnotesize
	\caption{\textbf{Anomaly detection results with AUROC metric on MVTec-AD}. All methods are evaluated under the unified case, where the learned model is applied to detect anomalies across all categories \textit{without} fine-tuning.}
	\vspace{-0.2cm}
	
	\resizebox{\textwidth}{!}{
		\begin{tabular}{c|c|*{10}{>{\centering\arraybackslash}p{1cm}}|*{5}{>{\centering\arraybackslash}p{1cm}}|c}
			\toprule
			\multirow{2}{*}{Setting} &\multirow{2}{*}{Method} & \multicolumn{10}{c|}{Object} & \multicolumn{5}{c|}{Texture} & \multirow{2}{*}{Avg} \\
			& & Bottle & Cable & Capsule & Hznut & MtlNut & Pill & Screw & Tbrush & Trans & Zipper & Carpet & Grid & Leather & Tile & Wood &   \\ 
			\midrule
			\multirow{8}{*}{\rotatebox{90}{one-class setting}} & S-T~\cite{us} & 84.0 & 60.0 & 57.6 & 95.8 & 62.7 & 56.1 & 66.9 & 57.8 & 61.0 & 78.6 & 86.6 & 69.2 & 97.2 & 93.7 & 90.6 & 74.5  \\
			&PSVDD~\cite{psvdd} & 85.5 & 64.4 & 61.3 & 83.9 & 80.9 & 89.4 & 80.9 & 99.4 & 77.5 & 77.8 & 63.3 & 66.0 & 60.8 & 88.3 & 72.1 & 76.8  \\ 
			&PaDiM~\cite{padim} & 97.9 & 70.9 & 73.4 & 85.5 & 88.0 & 68.8 & 56.9 & 95.3 & 86.6 & 79.7 & 93.8 & 73.9 & 99.9 & 93.3 & 98.4 & 84.2  \\ 
			&CutPaste~\cite{cutpaste} & 67.9 & 69.2 & 63.0 & 80.9 & 60.0 & 71.4 & 85.2 & 63.9 & 57.9 & 93.5 & 93.6 & 93.2 & 93.4 & 88.6 & 80.4 & 77.5  \\ 
			&MKD~\cite{mkd} & 98.7 & 78.2 & 68.3 & 97.1 & 64.9 & 79.7 & 75.6 & 75.3 & 73.4 & 87.4 & 69.8 & 83.8 & 93.6 & 89.5 & 93.4 & 81.9  \\ 
			&DR{\AE}M~\cite{draem} & 97.5 & 57.8 & 65.3 & 93.7 & 72.8 & 82.2 & 92.0 & 90.6 & 74.8 & 98.8 & 98.0 & {99.3} & 98.7 & {99.8} & \textbf{99.8} & 88.1  \\ 
			&AST~\cite{ast} & 96.9 & 92.3 & 93.6 & 98.6 & 98.8 & 84.9 & 92.0 & 97.2 & 96.1 & 86.2 & 97.6 & 93.4 & 99.4 & 96.6 & 98.2 & 94.7 \\
			&SimpleNet~\cite{simplenet} & 97.6 & 90.9 & 67.2 & 99.0 & 97.2 & 71.7 & 50.2 & 90.0 & 93.6 & 98.4 & 95.7 & 58.3 & 97.1 & 99.2 & 99.4 & 87.0 \\
			\midrule
			\multirow{8}{*}{\rotatebox{90}{multi-class setting}} & DiAD~\cite{diad} & 99.7 & 94.8 & 89.0 & 99.5 & 99.1 & 95.7 & 90.7 & 99.7 & 99.8 & 95.1 & 99.4 & 98.5 & 99.8 & 96.8 & 99.7 & 97.2 \\
			&UniAD~\cite{uniad} & 99.7 & 95.2 & 86.9 & 99.8 & 99.2 & 93.7 & 87.5 & 94.2 & 99.8 & 95.8 & 99.8 & 98.2 & \textbf{100} & 99.3 & 98.6 & 96.5  \\ 
			&UniAD(28$\times$28) & 99.8 & \textbf{96.4} & 83.2 & 99.9 & 98.2 & 82.1 & 90.6 & 93.3 & 99.8 & 94.1 & {99.9} & 93.6 & \textbf{100} & 99.1 & 98.2 & 95.2  \\ 
			&UniAD(56$\times$56) & 99.5 & 95.7 & 82.3 & 99.8 & 98.4 & 67.5 & 86.7 & 93.3 & 99.8 & 92.3 & \textbf{100} &\textbf{99.7} & \textbf{100} & 98.0 & 97.9 & 94.1  \\ 
			&UniAD+VAE & 99.7 & 95.9 & 86.7 & 99.8 & 98.8 & 93.3 & 87.8 & 93.6 & 99.7 & 96.0 &99.7 & 98.1 & \textbf{100} & 99.4 & 98.5 & 96.5 \\
			&UniAD+GAN & 99.5 & 94.6 & 79.4 & 99.2 & 99.0 & 73.8 & 86.0 & 89.7 & \textbf{99.9} & 91.3 & 99.9 & 98.7 & \textbf{100} & 97.9 & 97.7 & 93.8 \\
			&UniAD+UNet & 99.8 & 94.6 & 86.7 & 99.9 & 99.3 & 94.8 & 86.8 & 94.7 & 99.8 & 95.0 & 99.8 & 97.5 & \textbf{100} & 99.2 & 98.4 & 96.4 \\
			\cmidrule{2-18}
			&Ours & \textbf{100} & 94.7 & \textbf{94.0} & \textbf{100} & \textbf{99.9} & \textbf{96.9} & \textbf{95.6} & \textbf{100} & 99.8 & \textbf{99.7} & {98.5} &  {99.6} & \textbf{100} & \textbf{99.9} & {98.7} & \textbf{98.5} \\ 
			\bottomrule
	\end{tabular}}
	\vspace{-0.3cm}
		\label{tab:mvtec_det}

\end{table*}
	
	\subsectionvspace
	\subsection{Implementation Details}\label{subsec:details}
	All images used in our approach are resized to 224$\times$224 pixels and no augmented data involved. For the base model, we used a ImageNet~\cite{imagenet} pre-trained EfficientNet-b4~\cite{efficientnet} as a feature extractor. Features from stages 1 to 4 are selected, resized to 14$\times$14, and concatenated along channel dimension to form a multi-scale feature map $F_{\text{in}}$. 
	For our diffusion refinement branch, we use the same U-Net~\cite{unet} architecture as backend model from~\cite{diffusionbeatgans} and add an extra input as the model condition for noise prediction.
	During inference, we used the DDIM~\cite{ddim} method for deterministic sampling. 
	For the selection of the time period, the initial index of reverse diffusion is mainly considered to ensure that the samples contain low-frequency information; while the number of equally spaced samples to ensure the output heatmaps are accurate and stable.
	As a result, we empirically used timesteps of $t\!\in\!\{250,200,150,100,50,0\}$. Besides, we applied inpainting grid size $c\!\in\!\{1,16,32\}$, scale size $l\!\in\!\{1,2,4,8\}$, and $\gamma\!=\!0.9$, $n_s\!=\!2$, $m\!=\!41$.
	Both the base model and back-end model of diffusion are trained from the scratch by AdamW~\cite{adamw} with weight decay $1 \!\times\!10^{-4}$. All experimental results are presented as the average of five trials.

\begin{table*}[t]
	\setlength\tabcolsep{2.0pt}
	\centering
	\footnotesize

	\caption{\textbf{Anomaly localization results with AUROC metric on MVTec-AD}. All methods are evaluated under the unified case, where the learned model is applied to detect anomalies across all categories \textit{without} fine-tuning.}
	\vspace{-0.2cm}
	\resizebox{\textwidth}{!}{
		\begin{tabular}{c|c|*{10}{>{\centering\arraybackslash}p{1cm}}|*{5}{>{\centering\arraybackslash}p{1cm}}|c}
			\toprule
			\multirow{2}{*}{Setting} & \multirow{2}{*}{Method} & \multicolumn{10}{c|}{Object} & \multicolumn{5}{c|}{Texture} & \multirow{2}{*}{Avg} \\
			& & Bottle & Cable & Capsule & Hznut & MtlNut & Pill & Screw & Tbrush & Trans & Zipper & Carpet & Grid & Leather & Tile & Wood &   \\ 
			\midrule
			\multirow{8}{*}{\rotatebox{90}{one-class setting}} & S-T~\cite{us} & 67.9 & 78.3 & 85.5 & 93.7 & 76.6 & 80.3 & 90.8 & 86.9 & 68.3 & 84.2 & 88.7 & 64.5 & 95.4 & 82.7 & 83.3 & 81.8  \\ 
			&PSVDD~\cite{psvdd} & 86.7 & 62.2 & 83.1 & 97.4 & 96.0 & 96.5 & 74.3 & 98.0 & 78.5 & 95.1 & 78.6 & 70.8 & 93.5 & 92.1 & 80.7 & 85.6  \\ 
			&PaDiM~\cite{padim} & 96.1 & 81.0 & 96.9 & 96.3 & 84.8 & 87.7 & 94.1 & 95.6 & 92.3 & 94.8 & 97.6 & 71.0 & 84.8 & 80.5 & 89.1 & 89.5  \\ 
			&FCDD~\cite{fcdd} & 56.0 & 64.1 & 67.6 & 79.3 & 57.5 & 65.9 & 67.2 & 60.8 & 54.2 & 63.0 & 68.6 & 65.8 & 66.3 & 59.3 & 53.3 & 63.3  \\ 
			&MKD~\cite{mkd} & 91.8 & 89.3 & 88.3 & 91.2 & 64.2 & 69.7 & 92.1 & 88.9 & 71.7 & 86.1 & 95.5 & 82.3 & 96.7 & 85.3 & 80.5 & 84.9  \\ 
			&DR{\AE}M~\cite{draem} & 87.6 & 71.3 & 50.5 & 96.9 & 62.2 & 94.4 & 95.5 & 97.7 & 64.5 & {98.3} & {98.6} & \textbf{98.7} & 97.3 & \textbf{98.0} & \textbf{96.0} & 87.2  \\ 
			&AST~\cite{ast}  & 76.3 & 91.3 & 92.6 & 92.2 & 87.7 & 67.4 & 82.5 & 95.5 & 93.0 & 79.5 & 94.8 & 85.7 & 88.6 & 88.1 & 77.0 & 86.1 \\
			&SimpleNet~\cite{simplenet} & 94.3 & 91.3 & 95.3 & 95.0 & 91.0 & 88.9 & 93.0 & 95.2 & 91.0 & 96.3 & 97.6 & 49.5 & 97.1 & 93.6 & 88.8 & 90.5 \\
			\midrule
			\multirow{8}{*}{\rotatebox{90}{multi-class setting}} & DiAD~\cite{diad} & 98.4 & 96.8 & 97.1 & 98.3 & 97.3 & 95.7 & 97.9 & \textbf{99.0} & 95.1 & 96.2 & 98.6 & 96.6 & 98.8 & 92.4 & 93.3 & 96.8 \\
			&UniAD~\cite{uniad} & 98.1 & \textbf{97.3} & 98.5 & 98.1 & 94.8 & 95.0 & 98.3 & 98.4 & 97.9 & 96.8 & 98.5 & 96.5 & 98.8 & 91.8 & 93.2 & 96.8  \\ 
			&UniAD(28$\times$28) & 97.9 & 97.0 & 98.2 & 98.4 & 94.4 & 90.7 & 98.7 & 98.5 & 96.1 & 96.7 & 98.5 & 86.5 & 99.1 & 89.3 & 93.5 & 95.6  \\ 
			&UniAD(56$\times$56) & 97.6 & 96.6 & 97.9 & 98.5 & 94.0 & 90.0 & 98.1 & 98.2 & 96.8 & 96.1 & \textbf{98.7}  & 94.7 & \textbf{99.2} & 88.9 & 93.5 & 95.9  \\ 
			&UniAD+VAE & 98.0 & \textbf{97.3} & 98.4 & 98.2 & 93.5 & 94.9 & 98.4 & 98.4 & 98.2 & 96.5 & 98.5 & 96.5 & 98.9 & 92.0 & 93.3 & 96.7 \\
			&UniAD+GAN & 98.2 & 96.7 & 97.0 & 98.1 & 94.0 & 90.3 & 97.4 & 98.3 & 98.1 & 95.2 & 98.5 & 94.7 & 99.0 & 90.1 & 93.4 & 95.9 \\
			&UniAD+UNet & 98.1 & \textbf{97.3} & 98.5 & 98.0 & 93.2 & 95.2 & 98.3 & 98.4 & \textbf{98.4} & 96.4 & 98.5 & 96.4 & 98.8 & 91.9 & 93.3 & 96.7 \\
			\cmidrule{2-18}
			&Ours & \textbf{98.5} & {97.2} & \textbf{99.0} & \textbf{98.8} & \textbf{97.5} & \textbf{98.3} & \textbf{99.5} & 98.9 & 97.6 & \textbf{98.9} & \textbf{98.7} & \textbf{98.7} & \textbf{99.2} & 95.0 & 95.8 & \textbf{98.1} \\ 
			\bottomrule
	\end{tabular}}
	\vspace{-0.3cm}
		\label{tab:mvtec_loc}

\end{table*}

	\beforesectionvspace
	\section{Experiment}\label{sec:exp}
	\aftersectionvspace
	
	\subsection{Datasets and Metrics} \label{subsec:dataset}
	
	\textbf{Datasets}. \ul{MVTec-AD}~\cite{mvtec} includes 15 sub-categories and total 5,354 images, where 3,629 images are train images which are all normal, and 1,725 test images consist of both normal and anomalous images with ground-truth mask. Anomaly images are categorized with various kinds of defects. 
	\ul{BeanTechAD}~\cite{vtadl} is also a industrial anomaly detection dataset with 3 sub-categories and total 2,830 real-world images, among which 1,800 images are for training.
	
	\noindent\textbf{Metrics}. Following prior works~\cite{mvtec,us,draem}, the Area Under the Receiver Operating Curve (AUROC) is used as the evaluation metric for both anomaly detection and localization.
	
	\vspace{-0.1cm}
	\subsectionvspace
	\subsection{Comparisons to State-of-the-Art Methods}\label{subsec:exp}
	
	\textbf{Baselines}. Our approach is compared with baselines including: S-T~\cite{us}, PaDiM~\cite{padim}, PSVDD~\cite{psvdd}, MKD~\cite{mkd}, DR{\AE}M~\cite{draem}, AST~\cite{ast}, SimpleNet~\cite{simplenet}. 
		Noted that we only selected methods that can smoothly port to a multi-class setting for fair comparison, excluding those requiring significant network structure modifications to achieve acceptable performance. 
		All of these one-model-per-category methods above are compatible with multi-class inputs, and the records in the table are the performance of the unified model.
		Besides, we drawn comparisons with DiAD~\cite{diad}, UniAD~\cite{uniad}, originally designed for multi-class inputs.
		Due to the limited number of multi-class methods, we also compared with several custom variants of UniAD~\cite{uniad}. These include two variants with higher-resolution for feature reconstruction (28$\times$28, 56$\times$56, respectively), and three variants that incorporate additional original resolution image reconstruction branches to assist in the base heatmap fusion (vanilla VAE~\cite{vae}, vanilla GAN~\cite{gan}, U-Net~\cite{unet}, respectively). 
	All competitors are run with publicly available implementations.
	
	\noindent\textbf{Results of Anomaly Detection on MVTec-AD}
	are shown in \cref{tab:mvtec_det}. 
	Compared to most original one-model–per-category methods, our approach shows significant improvement (over 7.6\%) due to dual conditioning, which provides category-awareness and implicit classification capabilities during anomaly detection, without the need for ground-truth supervision. 
	Against one-model–all-categories methods, our proposed framework excels in most categories, surpassing UniAD~\cite{uniad} by 2.0\%, indicating our superiority. 
	Particularly in categories with more detail defects, such as screws or capsules, our method's advantage is even more pronounced (4.9\%$\sim$8.1\%).
	Compared to DiAD, which employs the LDM model and a denoising scheme, along with autoencoders for encoding and decoding, our approach improves by 1.3\% since the method still requires projecting original samples to embedding features, resulting in inevitable information loss and suboptimal results.
	Compared to various UniAD variants, we found that merely increasing the resolution of feature vectors doesn't enhance anomaly detection but weakens the Transformer's ability to correlate with context. With a growing number of feature tokens and limited training samples, the model struggles to find correlations between features, leading to overfitting and deteriorating performance. Variants integrating with VAE or GAN can reconstruct high-resolution images from latent codes but still suffer from spatial information loss or training instability. 
	The direct combination of U-Net is also not effective in improving the performance, since the U-Net image reconstruction needs to trade the image resolution (e.g. pooling and strided convolutions) for semantic information.
	In contrast, although the back-end model of our method is still U-Net, the final performance is improved by 1.4\% because the diffusion refinement always operates on the original resolution and the introduced inpainting effectively avoids ``identical shortcuts''.
	

	\noindent\textbf{Results of Anomaly Localization on MVTec-AD}
	are illustrated in \cref{tab:mvtec_loc}. 
	The improvements in anomaly localization are relatively smaller but still noteworthy, with a 2.7\% increase in metal nuts and a 3.3\% improvement in pills. Furthermore, our approach outperforms variants with high-resolution feature reconstruction and methods incorporating VAE, GAN, and U-Net for high-resolution image reconstruction in anomaly localization. These slight yet stable improvements emphasize our method's effectiveness, primarily attributed to the high-frequency refinement of image reconstruction by our modified diffusion.
	Remarkably, while previous methods have never reached 97\% AUROC, our method achieves state-of-the-art performance at 98.1\% in this field.

\begin{table}[t]
	\setlength\tabcolsep{3.0pt}
	\centering
	\footnotesize
	\caption{\textbf{Anomaly detection results with Image-wise AUROC (I-AU) and Pixel-wise AUROC (P-AU) on BeanTechAD} under the unified case.}
	\vspace{-0.2cm}
 	\footnotesize
		\begin{tabular}{c|cc|cc|cc|cc}
			\toprule
			\multirow{2}{*}{Method} & \multicolumn{2}{c|}{1}& \multicolumn{2}{c|}{2} & \multicolumn{2}{c|}{3} &\multicolumn{2}{c}{ Avg }\\
			&I-AU	&P-AU	&I-AU	&P-AU	&I-AU	&P-AU	&I-AU	&P-AU\\
			\midrule
   			DiAD & 95.91	&96.94	&80.56	&95.51	&90.98	&97.44	&89.15	&96.63 \\
			UniAD & 99.51	&97.00	&76.96	&95.14	&98.83	&98.63	&91.77	&96.92 \\
			UniAD(56$\times$56) &99.12	&97.55	&78.10	&95.06	&98.84	&98.70	&92.02	&97.10 \\
			UniAD+VAE &99.77	&96.92	&77.33	&94.90	&99.81	 &\textbf{99.68} 	&92.30	&97.17 \\
			UniAD+GAN & 98.15	&96.95	&77.95&	94.69	  &\textbf{99.82} &	99.57	&91.97&	97.07 \\
			UniAD+UNet &97.95	&97.02	&78.78	&94.81	&98.97	&99.46	&91.90	&97.13 \\
			Ours & \textbf{99.81} 	& \textbf{98.53} 	 &\textbf{82.07} 	& \textbf{96.75} 	 &{98.88} 	&99.50	& \textbf{93.59}	 &\textbf{98.26} \\
			\bottomrule
		\end{tabular}
	\vspace{-0.3cm}
 	\label{tab:btad}

\end{table}

\begin{table}[t]
	\centering
	\setlength\tabcolsep{3.0pt}
	\caption{\textbf{Ablation studies with AUROC metric on MVTec-AD} under the unified case.}
	\vspace{-0.2cm}
	\footnotesize
		\begin{tabular}{l|ccccccc}
			\toprule
			Multi-scale in base heatmap & $\times$ & $\times$ & $\times$ & $\times$ & $\times$ & \checkmark & $\times$ \\
			Spatio-temporal fusion & $\times$ & $\times$  & \checkmark & \checkmark & $\times$ & \checkmark & \checkmark \\  
			Model conditioning & $\times$ & \checkmark & $\times$ & \checkmark  & \checkmark & \checkmark & \checkmark \\
			Test-time conditioning & \checkmark & $\times$  & \checkmark & $\times$ & \checkmark & \checkmark & \checkmark \\ \midrule
			Image-wise AUROC  & 95.87 & 96.60 & 97.74 & 98.00 & 97.80 & 97.56 & \textbf{98.45} \\
			Pixel-wise AUROC & 96.43 & 97.15 & 98.02  & 97.82  & 97.98  & 97.69  & \textbf{98.05} \\
			\bottomrule
		\end{tabular}
		\vspace{-0.2cm}
  	\label{tab:ablation}

\end{table}

	
	\noindent\textbf{Results of Anomaly Detection and Localization on BeanTechAD} are reported in \cref{tab:btad}. The high-resolution variants of UniAD do not achieve an advantage in this dataset and even show performance degradation. In contrast,
	we achieve {93.59 image-wise AUROC} and {98.26 pixel-wise AUROC} which still stably outperform the average performance on UniAD by $\approx$2.0\%. Notably, our method also surpasses DiAD's experimental data, further demonstrating its effectiveness. In particular, samples of the 2 category have a relatively lower original performance and our \method get a larger boost, even outperforming the one-model–per-category methods~\cite{vtadl,panda,padim}, indicating the efficacy of our method. 
	
	
	\subsectionvspace
	\subsection{Ablation Studies}\label{subsec:ablation}
	To validate the effectiveness of our proposed modules, we conducted comprehensive ablation studies on the MVTec-AD dataset. As presented in \cref{tab:ablation}, the results provide a comparative analysis of the performance of different algorithm variants, evaluated by adding or removing individual modules.
	Compared with UniAD's performance, i.e. the baseline performance,
	we observe that simply applying test-time conditioning and spatio-temporal fusion can enhance the model's detection and localization by 1.2\%. Including the model condition further boosts the performance by 2.0\%. 
	However, integrating a multi-scale algorithm into the base heatmap lead to a 1.0\% decline in results. Therefore, we opted not to incorporate the multi-scale algorithm into the base heatmap in our final implementation. The ablation results highlight the effective contribution of each proposed module in our method.
	


\begin{figure*}[t]
\begin{minipage}[t]{0.19\textwidth}
\centering
\begin{tikzpicture}[scale=0.425] 
\begin{axis}[
    width=5.75cm,
    height=5.5cm,    
    ymin=97.2, ymax=98.5,
    xlabel=(a) timestep numbers $n_t$,
    ylabel=AUROC, 
    tick align=outside, 
    xmajorgrids=true,
    ymajorgrids=true,
    legend pos=south east,
    ]
\addplot[smooth,mark=*,blue] plot coordinates { 
    (4,97.4968)(8,98.0969)(12,98.0819)(16,98.0557)(20,98.0536)(24,98.0344)(28,98.0352)(32,98.0197)(36,98.025)(40,98.0152)(60,98.0255)(80,98.0258)(100,98.0298)(120,98.0298)
};
\addlegendentry{image-wise}
\addplot[smooth,mark=triangle,cyan] plot coordinates {
    (4,97.2876)(8,97.8412)(12,97.8478)(16,97.8485)(20,97.8501)(24,97.8488)(28,97.848)(32,97.8495)(36,97.8498)(40,97.8497)(60,97.8505)(80,97.8504)(100,97.8512)(120,97.8519)
    };
\addlegendentry{pixel-wise}
\end{axis}
\end{tikzpicture}
\end{minipage}
\begin{minipage}[t]{0.19\textwidth}
\centering
\begin{tikzpicture}[scale=0.425] 
\begin{axis}[
    width=6cm,
    height=5.5cm,    
    ymin=97.2, ymax=98.5,
    xlabel=(b) timestep initial index $t_{n_t}$,
    tick align=outside, 
    legend pos=north east,
    xmajorgrids=true,
    ymajorgrids=true,
    grid style=dashed,
    ]
\addplot[smooth,mark=*,blue] plot coordinates { 
    (50,97.4968)	(100,98.0049)	(150,98.0484)	(200,98.0695)	(250,98.0536)	(300,98.0257)	(350,97.9504)	(400,97.8324)	(450,97.5997)	(500,97.412)	
};
\addlegendentry{image-wise}
\addplot[smooth,mark=triangle,cyan] plot coordinates {
    (50,97.2876)	(100,97.7932)	(150,97.834)	(200,97.8467)	(250,97.8501)	(300,97.8441)	(350,97.8336)	(400,97.8188)	(450,97.7966)	(500,97.7624)	
};
\addlegendentry{pixel-wise}
\end{axis}
\end{tikzpicture}
\end{minipage}
\begin{minipage}[t]{0.19\textwidth}
\centering
\begin{tikzpicture}[scale=0.425] 
\begin{axis}[
    width=6cm,
    height=5.5cm,    
    ymin=97.2, ymax=98.5,
    xlabel=(c) inpainting ratio $n_s$, 
    tick align=outside, 
    legend pos=north east,
    xmajorgrids=true,
    ymajorgrids=true,
    grid style=dashed,
    ]
\addplot[smooth,mark=*,blue] plot coordinates { 
(2,98.362)	(3,98.2686)	(4,98.0536)	(5,97.7788)	(6,97.6428)	(7,97.5429)	(8,97.5321)
};
\addlegendentry{image-wise}
\addplot[smooth,mark=triangle,cyan] plot coordinates {
(2,97.9122)	(3,97.845)	(4,97.8501)	(5,97.8347)	(6,97.7828)	(7,97.7713)	(8,97.7575)
};
\addlegendentry{pixel-wise}
\end{axis}
\end{tikzpicture}
\end{minipage}
\begin{minipage}[t]{0.19\textwidth}
\centering
\begin{tikzpicture}[scale=0.425] 
\begin{axis}[
    width=6cm,
    height=5.5cm,
    ymin=97.3, ymax=98.5,
    xlabel=(d) mean filter size $m$ of fusion,
    tick align=outside, 
    legend pos=south east,
    xmajorgrids=true,
    ymajorgrids=true,
    grid style=dashed,
    ]
\addplot[smooth,mark=*,blue] plot coordinates { 
(11,98.0499)	(21,98.0639)	(31,98.0688)	(41,98.0536)	(51,98.0615)	(61,98.0405)	(71,98.0094)	(81,98.0078)};
\addlegendentry{image-wise}
\addplot[smooth,mark=triangle,cyan] plot coordinates {
(11,97.8761)	(21,97.879)	(31,97.8669)	(41,97.8501)	(51,97.8324)	(61,97.8154)	(71,97.7995)	(81,97.7849)};
\addlegendentry{pixel-wise}
\end{axis}
\end{tikzpicture}
\end{minipage}
\begin{minipage}[t]{0.19\textwidth}
\centering
\begin{tikzpicture}[scale=0.425] 
\begin{axis}[
    width=6cm,
    height=5.5cm,
    ymin=97.2, ymax=98.5,
    xlabel=(e) hyper-parameter $\gamma$ of fusion,
    tick align=outside, 
    legend pos=north west,
    xmajorgrids=true,
    ymajorgrids=true,
    grid style=dashed,
    ]
\addplot[smooth,mark=*,blue] plot coordinates { 
(0.1,97.2747)	(0.2,97.2963)	(0.3,97.318)	(0.4,97.384)	(0.5,97.4712)	(0.6,97.5352)	(0.7,97.717)	(0.8,97.8055)	(0.9,98.0536)
};
\addlegendentry{image-wise}
\addplot[smooth,mark=triangle,cyan] plot coordinates {
(0.1,97.2561)	(0.2,97.279)	(0.3,97.3069)	(0.4,97.3416)	(0.5,97.3861)	(0.6,97.445)	(0.7,97.5269)	(0.8,97.649)	(0.9,97.8501)
};
\addlegendentry{pixel-wise}
\end{axis}
\end{tikzpicture}
\end{minipage}
\vspace{-0.2cm}
\caption{Sensitivity analysis of various hyper-parameters of out \method method, including the number of timesteps $n_t$, the initial index of diffusion $t_{n_t}$, the number of disjoint sets $n_s$, the mean filter size $m$ and the hyper-parameter $\gamma$ of fusion.}
\label{fig:line}
\vspace{-0.4cm}
\end{figure*}
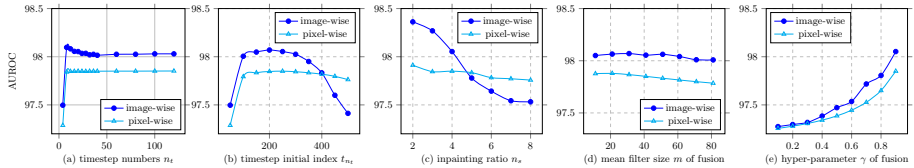

	\noindent\textbf{Sensitivity analysis} of various hyper-parameters is presented in \cref{fig:line}. It reveals that most of them are not sensitive, and our method can maintain high performance. 
	Specifically, this applies to the number of timesteps $n_t$ (as shown in \cref{fig:line}(a)), the initial index of diffusion $t_{n_t}$ (as shown in \cref{fig:line}(b)), and the mean filter size $m$ of fusion (as shown in \cref{fig:line}(e)).
	Since the base heatmap preserves the overall content of the reconstruction (low-frequency), only the high-frequency part is needed, i.e., only a few iterations (approximately 10) are required to achieve a stable boost instead of the original diffusion's 1000 sampling iterations.
	
	Following the previous definition~\cite{ddpm,ddim} (where step 0 represents the original image $x_0 (I_{ori})$ and step 1000 represents pure noise), we examined the initial index $t_{n_t}$ with the first 500 steps since only the high-frequency part is needed. The performance is illustrated in \cref{fig:line}(b), with the optimal results achieved at an initial index of 200-250.
	We also explored the number of disjoint sets $n_s$ (as shown in \cref{fig:line}(c)), determining that the best result can be obtained by simply splitting the original image $I_{ori}$ equally into known and unknown parts.
	The mean filter size $m$ in the fusion process was also analyzed (as shown in \cref{fig:line}(d)), revealing the model's insensitivity to this hyper-parameter.
	As displayed in \cref{fig:line}(e), setting the hyper-parameter $\gamma$ for fusion to 0.9 yields optimal performance. Notably, when $\gamma$ is set to 1.0, the model completely disregards the base heatmap, resulting in subpar performance as it only includes the high-frequency part. To better illustrate this hyper-parameter's trend, we excluded the case.
	
	\begin{figure*}[t]
		\centering
		\includegraphics[width=\linewidth]{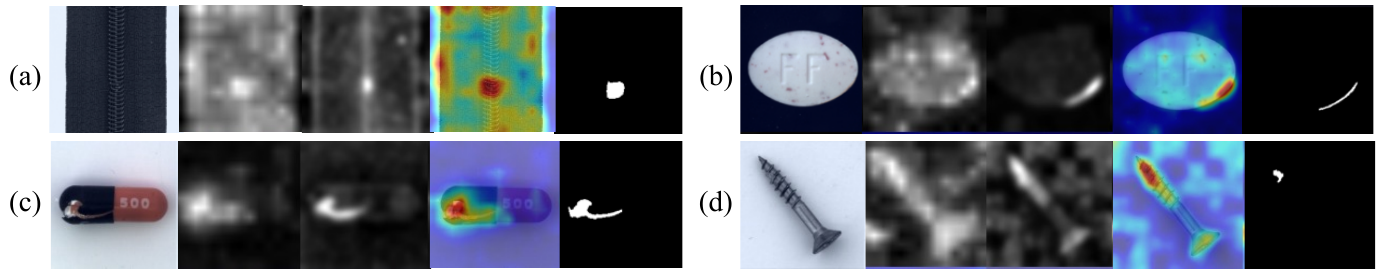}
		\vspace{-0.5cm}
		\caption{Examples of our proposed framework. Each case from left to right is the original image, the heatmap of the base model, the heatmap of the diffusion refinement, the final synthesized heatmap with the original image, and the ground-truth mask, respectively. Note that for illustration purposes, we only select $\mathcal{H}_{\text{diff}}$ at timestep $t\!=\!50$ as an example, while the output will fuse all generated heatmaps.}
		\label{fig:example}
		\vspace{-0.45cm}
	\end{figure*}
	
	\noindent\textbf{Visualization on MVTec-AD} 
	are shown in \cref{fig:example}. 
	The base model's heatmap is notably blurry, with indistinct boundaries and some activation cases present in the background region. 
	This is primarily due to its low spatial resolution, which hampers the performance. 
	However, our diffusion-based method, which predict noise and operate on the original image, naturally maintain the fineness of the original resolution.
	With the help of reverse diffusion process, the edges of the generated heatmap are clearer. 
	The final heatmaps, obtained through Spatio-temporal fusion, are comparatively pure. This is attributed to two factors: first, the diffusion process operates on the original resolution, resulting in less spatial information loss; second, we employ dual conditioning to boost reconstruction accuracy and use inpainting to further avoid ``identical shortcuts''.
	
	Histogram of the probability distribution and the comparison of the inference time of our \method are shown in the supplementary material, which shows that our \method exhibits a clearer boundary between normal and abnormal samples and that our method is more acceptable in industrial applications.
	\beforesectionvspace
	\section{Conclusion}
	\aftersectionvspace
	We discovered the low-resolution issue in previous reconstruction-based anomaly detection methods and proposed an noval framework by exploiting the diffusion model for refinement.
	The diffusion model is employed for the inpainting task to circumvent the ``identical shortcuts'' problem. To increase its sampling speed and takes full advantage of the reverse diffusion process, our method reconstructs only the high-frequency part to refine the original heatmap.
	To maintain accuracy and accommodate multi-class anomaly detection settings, we introduce dual conditioning for category-awareness and develop a Spatio-temporal fusion for smoother integration.
	Extensive experiments showcase the effectiveness of our method, and the contributions of each module are also carefully validated.
	

	%
	%
	\bibliographystyle{splncs04}
	\bibliography{main}
\end{document}


	\title{Supplementary Materials for Enhancing Multi-Class Anomaly Detection via Diffusion Refinement with Dual Conditioning} 
	
	
	\author{First Author\inst{1}\orcidlink{0000-1111-2222-3333} \and
		Second Author\inst{2,3}\orcidlink{1111-2222-3333-4444} \and
		Third Author\inst{3}\orcidlink{2222--3333-4444-5555}}
	
	\authorrunning{F.~Author et al.}
	
	\institute{Princeton University, Princeton NJ 08544, USA \and
		Springer Heidelberg, Tiergartenstr.~17, 69121 Heidelberg, Germany
		\email{lncs@springer.com}\\
		\url{http://www.springer.com/gp/computer-science/lncs} \and
		ABC Institute, Rupert-Karls-University Heidelberg, Heidelberg, Germany\\
		\email{\{abc,lncs\}@uni-heidelberg.de}}
	
	\maketitle


\begin{figure}[ht]
	\centering
	\includegraphics[width=\linewidth, height=4.05cm]{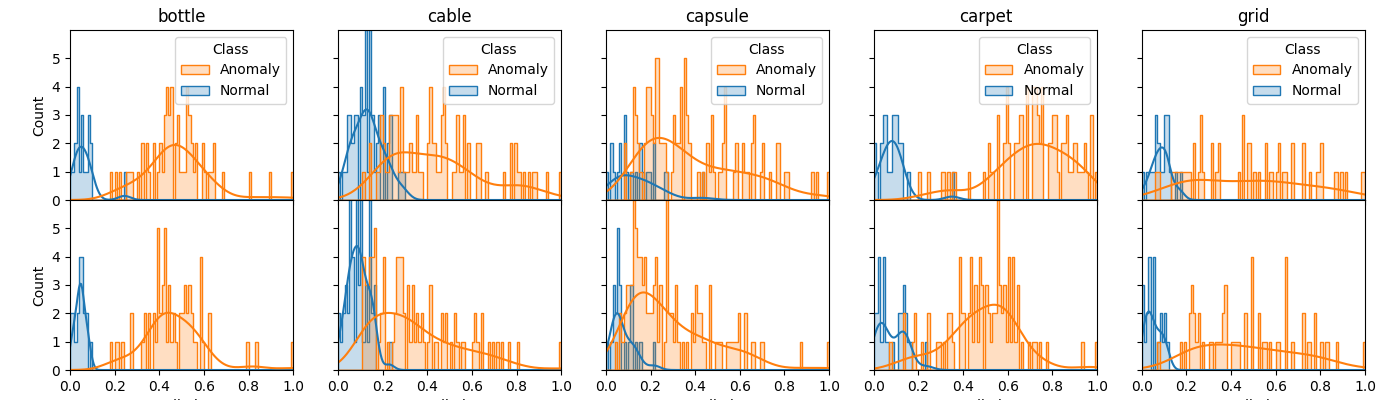}
	\includegraphics[width=\linewidth, height=4.05cm]{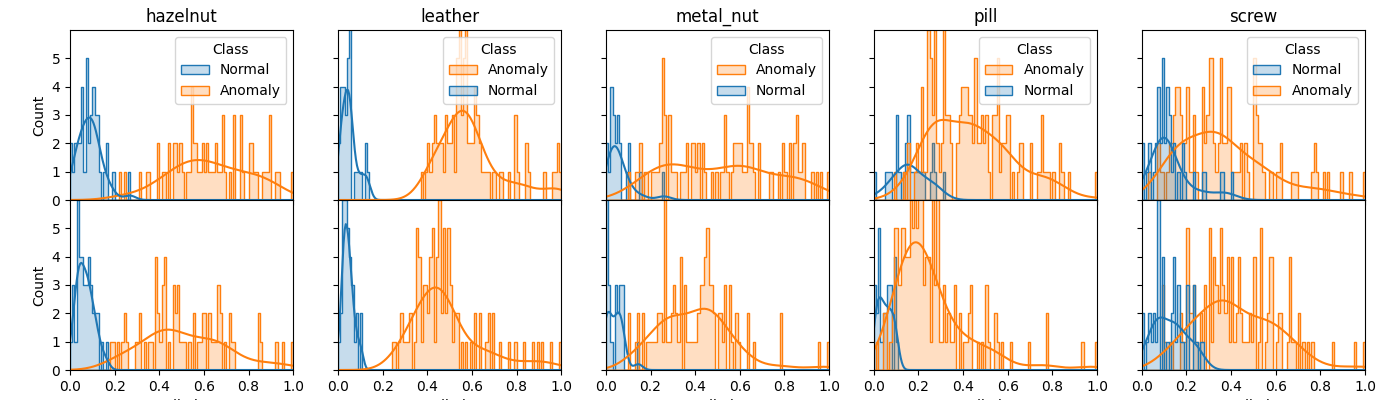}
	\includegraphics[width=\linewidth, height=4.05cm]{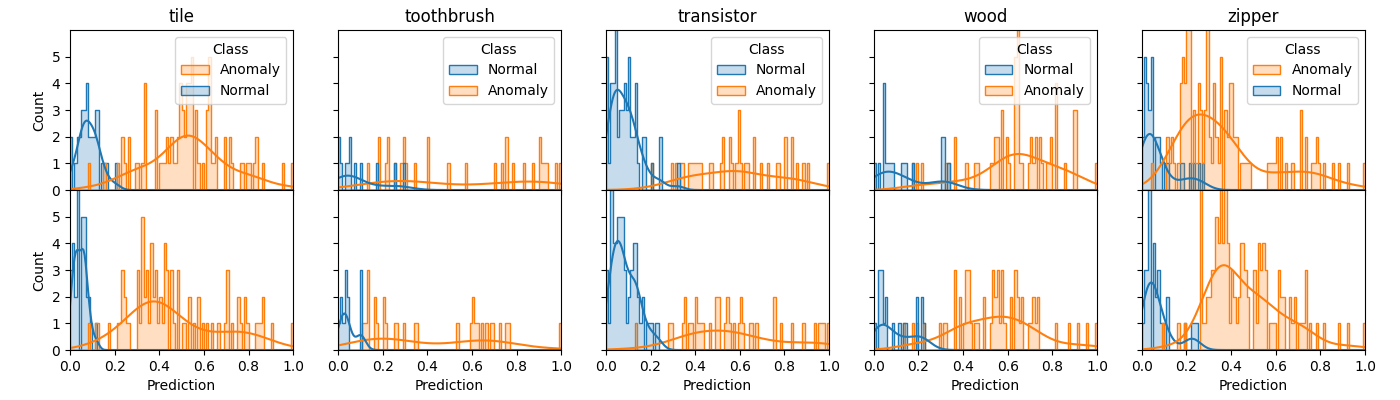}
	\caption{Comparison of histograms of normalized distributions of predicted probability for all subcategories on the MVTec-AD dataset. In which, the upper section is the histogram generated by UniAD [baseline], while the lower section is the histogram generated by our proposed \method.}
	\label{fig:hist}
\end{figure}

\section{Histogram on MVTec-AD}

\Cref{fig:hist} presents histograms illustrating the probability distribution for both our proposed \method method and the UniAD~\cite{uniad} baseline across various categories within the MVTec-AD dataset~\cite{mvtec}. It is evident from the histogram that our \method method delineates a more distinct boundary between normal and abnormal samples, particularly in categories such as Grid, Screw, Tile, Toothbrush, and Zipper. 
Based on the observations from the histograms, we find that the confusion area, i.e., the overlapping region between normal (blue) and abnormal (orange) samples, has decreased for most categories. Furthermore, the mean values of the normal and abnormal areas have become more distinct, indicating a clearer boundary between the two. This result suggests that our proposed model effectively reduces ambiguity and improves the differentiation between normal and abnormal samples across various categories.
This observation is in line with the superior performance exhibited in our quantitative analysis, which highlights the effectiveness of our method. 
\section{Performance and Inference Time}

It's worth mentioning that, although our \method does take longer in inference time compared to UniAD+GAN, as demonstrated in \Cref{tab:time}, it delivers a significant performance boost that is difficult to achieve when integrating UniAD with existing generative methods~\cite{gan,ddpm}.
In the table, we present the image-level AUROC and the inference time for a single image using various methods. It is evident that the time taken for inference using the method combined with the original diffusion is too long. 
Even though our method still has a gap compared to the method combined with GAN in terms of time, it remains quite acceptable for practical applications. 
More importantly, the considerable improvement in our image-level AUROC compared to other methods makes the tradeoff acceptable. 
As advancements in computational hardware continue, our \method holds the potential to become significantly faster, making it more fitting for future applications that are time-sensitive.

\begin{table}[h]
	\centering
 	\caption{Comparative analysis of image-level AUROC and time consumption on the MVTec-AD dataset: The comparison includes variants with vanilla GAN, variants with full diffusion, and our proposed \method method.}
	\begin{tabular}{l|c|r}
		\toprule
		\textbf{Method}                & \textbf{AUROC} & \textbf{Time Cost}        \\ 
		\midrule
		UniAD+GAN             & 93.8  & 0.293s      \\ 
		UniAD+Diffusion(Full) & 96.8  & 58min17s    \\ 
		Ours                  & 98.5  & 14.816s     \\
		\bottomrule
	\end{tabular}
	\label{tab:time}
\end{table}

\section{Extra Visualization on MVTec-AD}
\begin{figure}[ht]
	\centering
	\includegraphics[width=\linewidth]{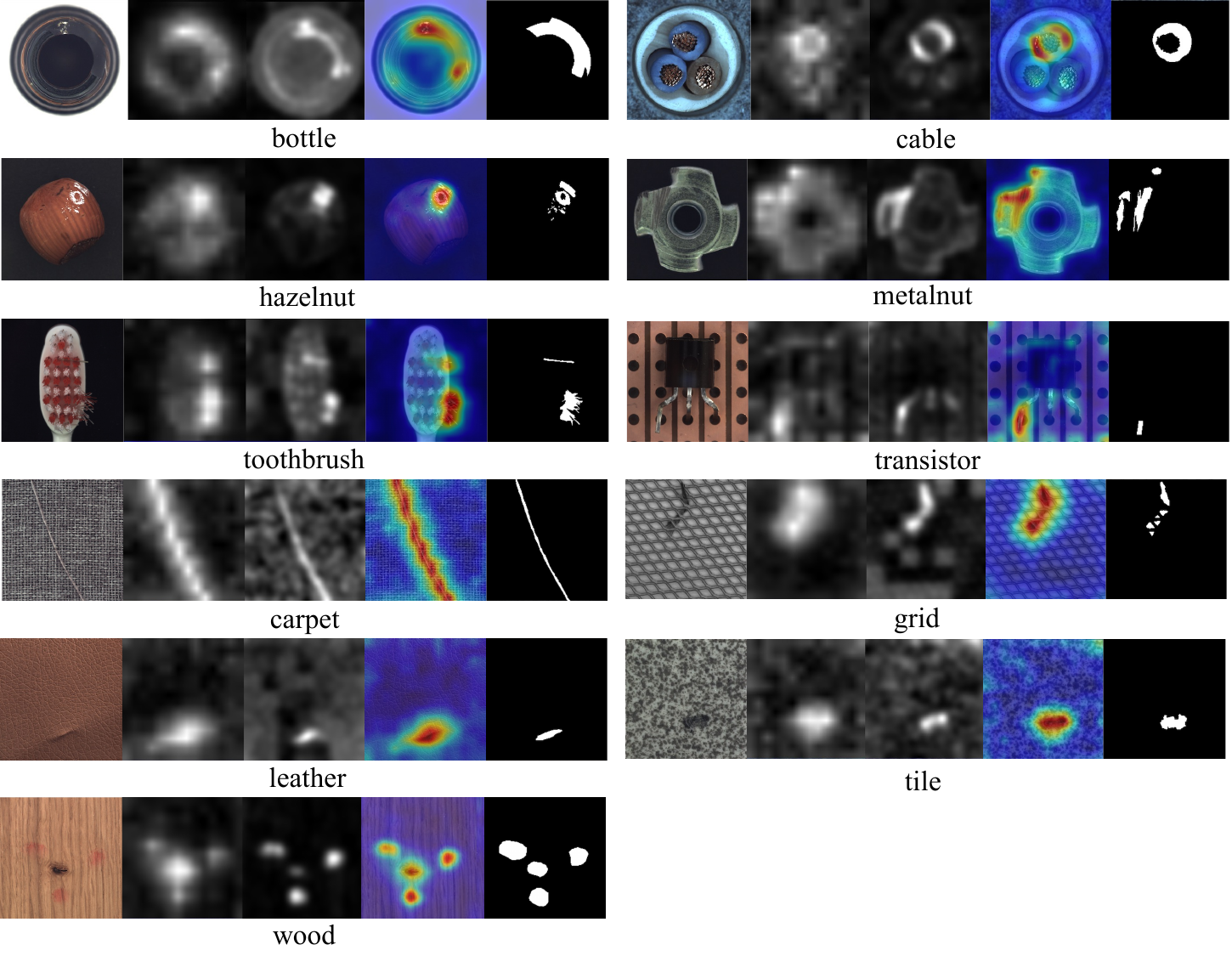}
	\caption{Extra examples of our proposed framework. Each case from left to right is the original image, the heatmap of the base model, the heatmap of the diffusion refinement, the final synthesized heatmap with the original image, and the ground truth mask, respectively.}
	\label{fig:extra}
\end{figure}

Here, we provide extra visualizations for more categories as shown in~\Cref{fig:extra}, in addition to the four categories (Zipper, Capsule, Pile, Screw) mentioned in the main paper. 

These visualizations further confirm the conclusions of our main paper.
Despite the base heatmap's blurry appearance due to its low resolution, our Diffusion Refinement method effectively enhances the boundary clarity in the generated heatmap, thereby improving the final anomaly detection results.
This improvement is particularly noticeable in categories such as Carpet, Grid, and Leather, which contain thin anomalies in their textures. Similarly, in categories such as Tile, Metalnut, Wood, and Cables, our method's ability to detect irregular defects in burrs and internal boundaries is significantly enhanced. 

 The visualizations not only supports the enhanced ability of our proposed model to differentiate between normal and abnormal samples, but also provides insights into the model's performance in various real-world scenarios. For example, in the Hazelnut category, our method can better detect dirt and contamination, but its performance is average in identifying dents. In the Wood category, it can more accurately recognize color inconsistencies, but its performance in detecting oxidation in the wood itself is average. In the Toothbrush category, our method excels at detecting bristle issues, such as uneven or missing bristles, but its performance in identifying foreign objects is average. These examples further emphasize the practical applicability of our proposed model in a diverse range of anomaly detection tasks, while also highlighting areas for potential improvement.
 
\section{Experiments on VisA Dataset}
In this section, we further describe the experiments conducted on the latest VisA dataset. The VisA dataset~\cite{visa} comprises a total of 10,821 high-resolution images, including 9,621 normal images and 1,200 abnormal images, covering 78 types of anomalies. The VisA dataset is composed of 12 subsets, each corresponding to a different object. The 12 objects can be categorized into three different types: complex structures, multiple instances, and single instance. We report the comparative results of typical methods including DR{\AE}M~\cite{draem}, UniAD~\cite{uniad}, DiAD~\cite{diad}, and our approach. 
As can be seen from the table~\Cref{tab:visa_det} and~\Cref{tab:visa_loc}, 
our proposed \method still exhibits significant improvements over other methods in both anomaly detection and localization tasks on the VisA dataset, further validating the effectiveness of the our method.

\begin{table*}[t]
	\setlength\tabcolsep{2.0pt}
	\centering
	\small
	\caption{\textbf{Anomaly detection results with AUROC metric on VisA}. All methods are evaluated under the unified case, where the learned model is applied to detect anomalies across all categories \textit{without} fine-tuning.}
	
	\resizebox{\textwidth}{!}{
		\begin{tabular}{c|*{4}{>{\centering\arraybackslash}p{1cm}}|*{4}{>{\centering\arraybackslash}p{1cm}}|*{4}{>{\centering\arraybackslash}p{1cm}}|*{1}{>{\centering\arraybackslash}p{1cm}}}
			\toprule
			\multirow{2}{*}{Method} & \multicolumn{4}{c|}{complex structures} & \multicolumn{4}{c|}{multiple instances} & \multicolumn{4}{c|}{single instance} & \multirow{2}{*}{Avg} \\
   
			  & pcb1 & pcb2 & pcb3 & pcb4 & mac1 & mac2 & caps & cand & cashew & gum & fryum & pipefry  \\ 
			\midrule
			DR{\AE}M~\cite{draem} & 71.9 & 78.4 & 76.6 & 97.3 & 69.8 & 59.4 & \textbf{83.4} & 69.3 & 81.7 & 93.7 & 89.1 & 82.8 & 79.1   \\
			UniAD~\cite{uniad} & 92.8 & 87.8 & 78.6 & 98.8 & 79.9 & 71.6 & 55.6 & 94.1 & \textbf{92.8} & 96.3 & 83.0 & 94.7 & 85.5  \\ 
			DiAD~\cite{diad} & 88.1 & 91.4 & \textbf{86.2} & \textbf{99.6} & 85.7 & 62.5 & 58.2 & 92.8 & 91.5 & \textbf{99.1} & \textbf{89.8} & \textbf{96.2} & 86.8 \\ 
			Ours & \textbf{94.0} & \textbf{91.8} & \textbf{86.2} & 98.3 & \textbf{90.0} & \textbf{81.8} & 70.1 & \textbf{95.9} & \textbf{92.8} & 97.8 & 88.2 & 96.0 &\textbf{90.2} \\ 
			\bottomrule
	\end{tabular}}
		\label{tab:visa_det}
\end{table*}

\begin{table*}[t]
	\setlength\tabcolsep{2.0pt}
	\centering
	\small
	\caption{\textbf{Anomaly localization results with AUROC metric on VisA}. All methods are evaluated under the unified case, where the learned model is applied to detect anomalies across all categories \textit{without} fine-tuning.}
	
	\resizebox{\textwidth}{!}{
		\begin{tabular}{c|*{4}{>{\centering\arraybackslash}p{1cm}}|*{4}{>{\centering\arraybackslash}p{1cm}}|*{4}{>{\centering\arraybackslash}p{1cm}}|*{1}{>{\centering\arraybackslash}p{1cm}}}
			\toprule
			\multirow{2}{*}{Method} & \multicolumn{4}{c|}{complex structures} & \multicolumn{4}{c|}{multiple instances} & \multicolumn{4}{c|}{single instance} & \multirow{2}{*}{Avg} \\
   
			  & pcb1 & pcb2 & pcb3 & pcb4 & mac1 & mac2 & caps & cand & cashew & chewgum & fryum & pipefry  \\ 
			\midrule
			DR{\AE}M~\cite{draem} & 94.6 & 92.3 & 90.8 & 94.4 & 95.0 & 94.6 & 97.1 & 82.2 & 80.7 & 91.0 & 92.4 & 91.1 & 91.3   \\
			UniAD~\cite{uniad} & 93.3 & 93.9 & \textbf{97.3} & 94.9 & 97.4 & 95.2 & 88.7 & \textbf{98.5} & \textbf{98.6} & \textbf{98.8} & 95.9 & 98.9 & 95.2   \\ 
			DiAD~\cite{diad} & \textbf{98.7} & 95.2 & 96.7 &\textbf{97.0} & 94.1 & 93.6 & \textbf{97.3} & 97.3 & 90.9 & 94.7 & \textbf{97.6} & \textbf{99.4} & 96.0 \\ 
			Ours & 98.3 & \textbf{96.6} & 97.1 & 96.7 & \textbf{98.1} & \textbf{96.6} & 97.1 & 98.0 & {97.9} & 98.0 & 96.7 &  98.2 & \textbf{97.4} \\ 
			\bottomrule
	\end{tabular}}
		\label{tab:visa_loc}
\end{table*}

	
	
	\bibliographystyle{splncs04}
	\bibliography{main}